\documentclass[10pt]{cai26}

\begin{document}
\def\conferenceyear{2026}
\volumeheader{39}{0}
\begin{center}

\title{FogTTA: Online Test-Time Adaptation for Robust Transformer-based Object Detection in Foggy Weather}
\maketitle

\thispagestyle{empty}
\pagenumbering{gobble}

\begin{tabular}{cc}
Ali Abedi\upstairs{\affilone,*}, Q.M. Jonathan Wu\upstairs{\affilone}, Ning Zhang\upstairs{\affilone}, Shiqi Tian\upstairs{\affiltwo}
\\[0.25ex]
{\small \upstairs{\affilone} University of Windsor} \\
{\small \upstairs{\affiltwo} University of Toronto Mississauga} \\
\end{tabular}
  
\emails{
  \upstairs{*} abedi3@uwindsor.ca 
}
\vspace*{0.1in}
\end{center}

\begin{abstract}
Object detection models for autonomous driving commonly experience substantial performance drops when deployed under adverse weather due to the domain shift between training data and real-world operating conditions. This degradation is especially evident when models trained on clear-weather images encounter foggy environments with reduced visibility and contrast. To address this challenge, we introduce FogTTA, an online test-time adaptation framework designed to improve the robustness of Transformer-based object detectors in fog. Using RF-DETR as the underlying object detector, FogTTA enables real-time adaptation to the streaming target domain without requiring source data or retraining. The framework follows a teacher–student design, where the deployed model serves as the teacher and generates pseudo labels from weakly augmented target inputs. These predictions are subsequently refined through non-maximum suppression and confidence filtering. The student model then learns from strongly augmented target sample using the Varifocal loss to mitigate pseudo-label noise. The teacher is updated via exponential moving averaging to ensure stable and continuous adaptation. Experiments show that FogTTA outperforms prior baselines, delivering improved detection accuracy and stability while maintaining real-time performance.
\end{abstract}

\copyrightnotice

\section{Introduction}

Object detection is a fundamental task in autonomous driving, where reliable perception under diverse environmental conditions is essential for safety~\cite{10.1561/0600000079}. Accurate detection and localization of surrounding objects are essential for safe autonomous navigation. Early detection models are based on convolutional neural networks (CNNs) and can be categorized into two-stage and one-stage approaches. Two-stage detectors, such as Faster R-CNN~\cite{NIPS2015_14bfa6bb}, achieve high accuracy through region proposal refinement but incur heavy computational costs. As real-time performance is critical for autonomous vehicles, one-stage detectors such as SSD~\cite{10.1007/978-3-319-46448-0_2} and YOLO~\cite{Redmon_2016_CVPR} were developed to deliver efficient detection through a single feed-forward pass. CNN-based methods rely on local receptive fields and thus struggle to model long-range dependencies, which limits their adaptability to adverse conditions~\cite{dosovitskiy2020image, 10.1007/978-3-030-58452-8_13}.

Transformer-based architectures, on the other hand, leverage self-attention to capture long-range dependencies and global context across entire images. Early models such as DETR~\cite{10.1007/978-3-030-58452-8_13} demonstrated strong accuracy but suffered from high computational overhead and slow convergence. Subsequent detectors, including Deformable DETR~\cite{zhu2020deformable}, LW-DETR~\cite{chen2024lwdetrtransformerreplacementyolo}, and RT-DETR~\cite{Zhao_2024_CVPR}, improved speed and efficiency while maintaining competitive accuracy. RF-DETR~\cite{rf-detr} integrates DINOv2 into LW-DETR, which makes it a robust and efficient detector. However, models trained under clear-weather conditions still experience substantial performance degradation when deployed in foggy environments due to the domain shift.

\begin{figure*}[t]
\centering
\begin{minipage}{\linewidth}
\includegraphics[width=\textwidth]{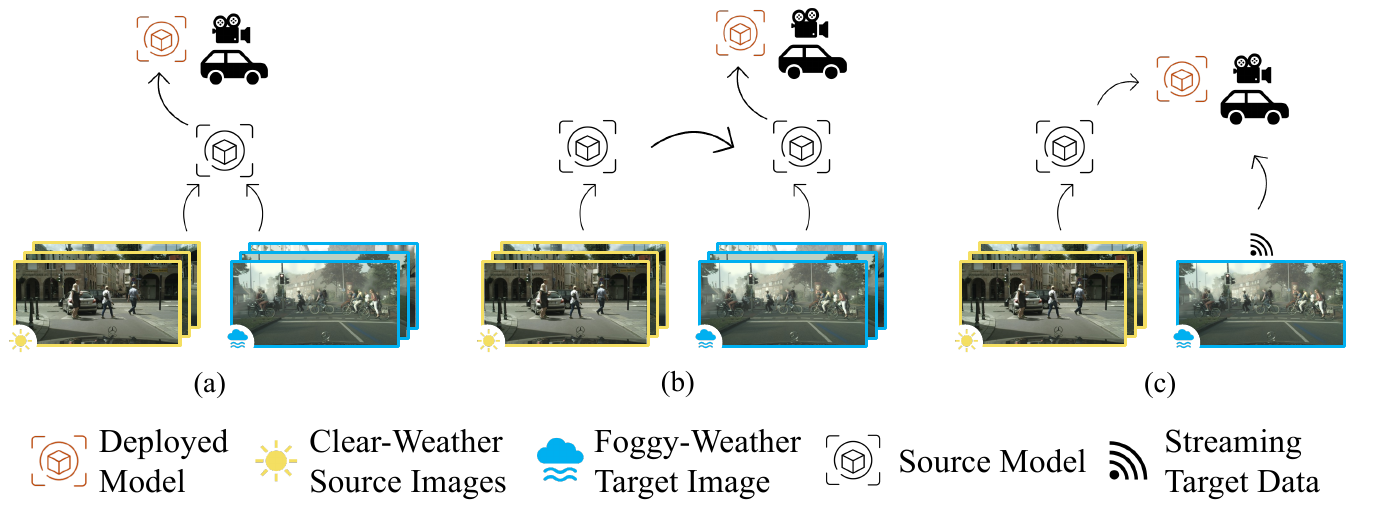}
\caption{Illustration of domain adaptation paradigms for object detection: (a) UDA uses labeled source and unlabeled target data jointly; (b) SFDA adapts a source-trained model without source access; (c) TTA adapts online to streaming target data during deployment.}
\label{fig:DA_Variants}
\end{minipage}
\end{figure*}

Domain shift refers to the discrepancy between the distributions of the training (source) and deployment (target) environments~\cite{10452835}. In autonomous driving, this shift is critical when clear-weather models face fog, where reduced visibility and contrast hinder object detection~\cite{10030451}. Unsupervised domain adaptation (UDA) methods aim to mitigate this issue by aligning labeled source and unlabeled target data. However, as shown in Figure~\ref{fig:DA_Variants}a, such methods~\cite{10030451, 10574400} assume access to both domains, which is impractical due to privacy and storage constraints. Source-free domain adaptation (SFDA)~\cite{11164935, he2025dual} addresses this limitation by adapting pre-trained models to unlabeled target data without accessing source samples (Figure~\ref{fig:DA_Variants}b), yet still requires offline retraining on the entire target set. In contrast, online test-time adaptation (TTA)~\cite{ZHANG2025105740, 10713112, VS_2023_WACV} enables real-time adaptation to a stream of unlabeled target inputs during deployment (Figure~\ref{fig:DA_Variants}c). Although more practical, TTA remains challenging due to the absence of supervision, the noise in pseudo labels, and the need to balance adaptation stability and responsiveness.

In this paper, we present FogTTA, an online test-time adaptation framework that enhances the robustness of Transformer-based object detectors under foggy conditions. Built upon RF-DETR~\cite{rf-detr}, FogTTA adapts continuously to unseen target data without requiring source access or retraining. It employs a teacher–student design, where the deployed model (teacher) generates and refines pseudo labels through confidence-based filtering and non-maximum suppression (NMS), while the student learns from strongly augmented target views using the Varifocal loss \cite{zhang2021varifocalnetiouawaredenseobject} to handle pseudo-label noise. The teacher is updated via exponential moving averaging (EMA) of the student’s parameters to ensure stable and progressive adaptation.  

The main contributions of this work are as follows:
\begin{itemize}
    \item We propose FogTTA, an online test-time adaptation framework that enhances the robustness of Transformer-based object detection under foggy weather conditions without requiring source data.  
    \item We utilized a pseudo-label refinement strategy that integrates NMS and the Varifocal loss to mitigate noise and uncertainty during adaptation.  
    \item We validate FogTTA through experiments, demonstrating improved detection accuracy in foggy conditions while maintaining real-time inference performance of RF-DETR.
\end{itemize}

The rest of this paper is organized as follows. Section~\ref{sec:related_work} reviews related work. Section~\ref{sec:method} presents the proposed methodology. Section~\ref{sec:experimental_settings} describes the experimental settings, and Section~\ref{sec:results} discusses the results. Finally, Section~\ref{sec:conclusion} concludes the paper and outlines future directions.

\section{Related Works}
\label{sec:related_work}
\subsection{Transformer-based Object Detectors}
The introduction of DETR \cite{10.1007/978-3-030-58452-8_13} marked a turning point in object detection. As the first Transformer-based detector, DETR differs from previous approaches by utilizing learnable object queries and an end-to-end pipeline that eliminates the need for post-processing. However, the original DETR design is not suitable for real-time applications due to its computational complexity and relatively high latency. Deformable DETR \cite{zhu2020deformable}, a follow-up work, adopts deformable attention to focus on key regions, accelerating training and reducing memory usage. To further improve real-time performance, RT-DETR \cite{Zhao_2024_CVPR} replaces the Transformer encoder with an efficient hybrid encoder to decouple the intra-scale interaction and cross-scale fusion of multi-scale features, improving performance while preserving the NMS-free design for faster inference. LW-DETR \cite{chen2024lwdetrtransformerreplacementyolo} uses a convolutional projector to connect a plain ViT encoder and a shallow DETR decoder, yielding a light-weight detector suitable for real-time applications. Building on LW-DETR and DINOv2, RF-DETR \cite{rf-detr} leverages neural architecture search to create a scheduler-free object detector that surpasses prior state-of-the-art approaches as well as real-time methods.

\subsection{Cross-weather domain adaptation}
To systematically study clear-to-foggy adaptation, Foggy Cityscapes~\cite{8917518} dataset synthesizes foggy scenes on Cityscapes~\cite{Cordts2016Cityscapes} dataset and is widely used as a benchmark for weather-robust object detection tasks. Early work on cross-weather domain adaptive object detection mainly follows the UDA setting. DA-Detect \cite{10030451} augments Faster RCNN with image-level and object-level domain classifiers plus an auxiliary domain to reduce the influence of domain shift. This framework is further extended \cite{10574400} to foggy and rainy conditions with a Dynamic Masking Process that exploits contextual information. However, these methods require access to both source and target data during training, thereby limiting their practicality in real-world deployment. To avoid accessing source data, DDT with AEMA \cite{he2025dual} introduces a Dual-Rate Dynamic Teacher (DDT) with Asynchronous EMA (AEMA) to balance fast adaptation to domain shift and long-term knowledge in a source-free setting. FRANCK \cite{11164935} shows a query-centric source-free framework for DETR with objectness-based sample reweighting module, a contrastive learning bank, and an uncertainty-weighted distillation. Nevertheless, these methods all rely on offline adaptation on a pre-collected target set, which limits their applicability in online test-time scenarios. 

\subsection{Test-time Domain Adaptation}
Recent work extends TTA and online adaptation ideas to object detection by adapting detectors during inference without source data. MemCLR \cite{VS_2023_WACV} introduced a unified teacher-student framework that adapts the target domain in both offline and online settings. MLFA \cite{10713112} aligns informative global-level and cluster-level features between source and target domains while minimizing target entropy for realistic TTA object detection. DDT \cite{ZHANG2025105740} introduce Dynamic Dual Teaching (DDT) scheme that combines a source detector and a vision language model teacher with dynamic prediction fusion and consistency regularization to achieve better generalization ability. However, existing TTA detectors still struggle with noisy pseudo labels and unstable adaptation under severe fog conditions. To address these limitations, we develop a transformer-based TTA framework that performs reliable adaptation without accessing source data.

\section{Proposed Method}
\label{sec:method}
\subsection{Problem Formulation}
Let the labeled source domain be denoted as $\mathcal{D}^{S} = \{(x_i^{S}, y_i^{S})\}_{i=1}^{N_S}$, drawn from the joint distribution $P_S(X, Y)$, where $x_i^{S}$ and $y_i^{S}$ represent the input image and its corresponding annotation. The source domain corresponds to clear-weather driving conditions. A model $f_{\theta}$ is first trained on the source domain by minimizing the supervised detection loss:
\begin{equation}
\min_{\theta} \mathcal{L}_{\text{src}}(f_{\theta}) = \frac{1}{N_S} \sum_{i=1}^{N_S} \ell(f_{\theta}(x_i^{S}), y_i^{S}),
\end{equation}
where $\ell(\cdot, \cdot)$ denotes the overall detection loss that combines classification and localization terms. 

At deployment, the pretrained model encounters a stream of unlabeled target samples $\mathcal{D}^{T} = \{x_j^{T}\}_{j=1}^{N_T}$, drawn from foggy-weather driving conditions following a distinct distribution $P_T(X)$ such that $P_T(X) \neq P_S(X)$. The goal of TTA is to adapt $f_{\theta}$ online to the target distribution without access to $\mathcal{D}^{S}$ or its labels. Formally, the adaptation objective is expressed as:
\begin{equation}
\min_{\theta} \mathbb{E}_{x^{T} \sim P_T(X)} \left[\, \ell(f_{\theta}(x^{T})) \,\right],
\end{equation}
where $\mathbb{E}_{x^{T} \sim P_T(X)}[\cdot]$ denotes the expected loss over target samples drawn from the target distribution, and $\ell(f_{\theta}(x^{T}))$ represents the unsupervised adaptation loss computed from each incoming target image.

\subsection{Model Overview}
\label{sec:overview}
\begin{figure*}[t]
\centering
\begin{minipage}{\linewidth}
\includegraphics[width=\textwidth]{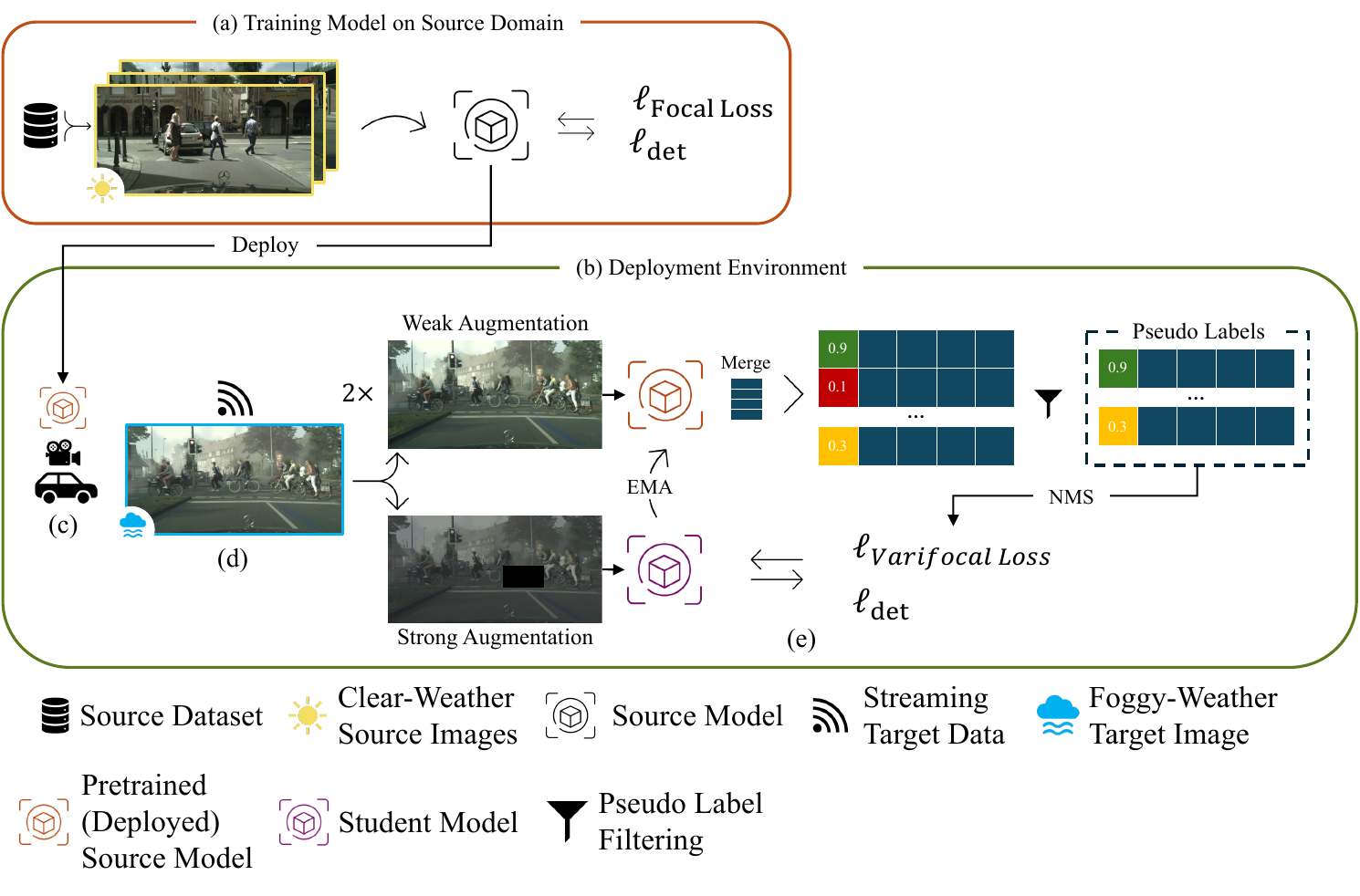}
\caption{Model overview of the proposed FogTTA framework. (a) The model is trained on the labeled source domain using focal and detection losses. (b) The trained model is then deployed on the vehicle. (c) During deployment, the vehicle encounters a continuous stream of target data with a distribution different from the source domain. (d) The incoming target data are processed through the TTA pipeline. (e) Weak and strong augmentations of the target images are generated, where the deployed model serves as the teacher and processes weakly augmented inputs to produce refined pseudo labels. NMS is applied to the results, and the student model, which receives the strongly augmented inputs, is trained on these pseudo labels using the Varifocal loss. The teacher model is then updated via EMA of the student’s parameters.}
\label{fig:model_overview}
\end{minipage}
\end{figure*}

Figure~\ref{fig:model_overview} illustrates the overall framework of our proposed TTA method for autonomous driving under foggy conditions. The process begins with a model $f_{\theta}$ pretrained on a labeled source dataset $\mathcal{D}^{S}$ that contains clear-weather driving scenes, as described in Section~\ref{sec:source_train}. During deployment, this pretrained model is installed on the vehicle and performs inference on a continuous stream of unlabeled target samples $\mathcal{D}^{T}$ captured under foggy-weather conditions. As the visual appearance of foggy scenes differs significantly from clear-weather training data, the model experiences performance degradation caused by the domain shift between $P_T(X)$ and $P_S(X)$.

To maintain robustness under such domain shifts, the model is continuously adapted to the incoming target data in an online manner without accessing the source dataset. This process follows a teacher–student framework, where the pretrained model $f_{\theta}$ serves as both the inference and teacher network, and a student model $f_{\phi}$, initialized with the teacher’s parameters, is used for adaptation. This architecture enables continuous online refinement as the environment evolves while ensuring that inference remains real-time through the efficiency of the RF-DETR backbone and is unaffected by the adaptation updates.

As detailed in Section~\ref{sec:tta}, for each target sample in the stream, the teacher generates predictions that are treated as supervision signals for the student. The student is optimized using the Varifocal loss~\cite{zhang2021varifocalnetiouawaredenseobject} (Section~\ref{sec:vertifocal_loss}) to minimize the discrepancy between its predictions and the teacher’s refined pseudo labels, thereby learning target-specific representations from unlabeled data. After each adaptation step, the teacher model is updated using an EMA of the student’s parameters to integrate the latest knowledge from the target domain. Consequently, the teacher model, which also serves as the deployed inference model, remains continuously adapted to the evolving target data stream while retaining the knowledge learned from the source domain.

\subsection{Training Source Model}
\label{sec:source_train}
The training objective of source $f_{\theta}$ on the labeled source dataset $\mathcal{D}^{S}$ follows the standard DETR \cite{10.1007/978-3-030-58452-8_13} formulation and is optimized using a combination of classification and localization losses. Specifically, the overall loss is defined as:
\begin{equation}
\mathcal{L}_{\text{src}} = \lambda_{\text{cls}} \mathcal{L}_{\text{cls}} + \lambda_{\text{box}} \mathcal{L}_{\text{box}} + \lambda_{\text{giou}} \mathcal{L}_{\text{giou}},
\label{eq:src_loss}
\end{equation}
where $\lambda_{\text{cls}}$, $\lambda_{\text{box}}$, and $\lambda_{\text{giou}}$ are weighting coefficients that balance the individual loss terms.

The classification loss $\mathcal{L}_{\text{cls}}$ is computed using the focal loss~\cite{lin2018focallossdenseobject}, which addresses the foreground–background imbalance commonly observed in dense detection tasks. It is formulated as:
\begin{equation}
\mathcal{L}_{\text{cls}} = - \alpha_t (1 - p_t)^{\gamma} \log(p_t),
\end{equation}
where $p_t$ represents the predicted probability for the target class, $\alpha_t$ is a balancing factor, and $\gamma$ is the focusing parameter that down-weights easy samples.

For localization, the bounding box regression loss $\mathcal{L}_{\text{box}}$ and the generalized IoU loss $\mathcal{L}_{\text{giou}}$ \cite{rezatofighi2019generalizedintersectionunionmetric} are employed to jointly optimize box position and shape alignment. The Hungarian matching algorithm \cite{10.1007/978-3-030-58452-8_13} is used to associate each prediction with a corresponding ground-truth box before loss computation.

\subsection{Test-Time Domain Adaptation}
\label{sec:tta}
For each incoming image $x_t$, the teacher network generates a set of object predictions
\[
\mathcal{P}_t = \{(b_i, p_i)\}_{i=1}^{N_t},
\]
where $b_i$ and $p_i$ represent the predicted bounding box and its associated class probability vector. To improve stability, the teacher performs two inference passes on weakly augmented versions of $x_t$, and the resulting predictions are merged and ranked according to their confidence scores.

\begin{figure}[htbp]
\centering
\includegraphics[width=0.5\textwidth]{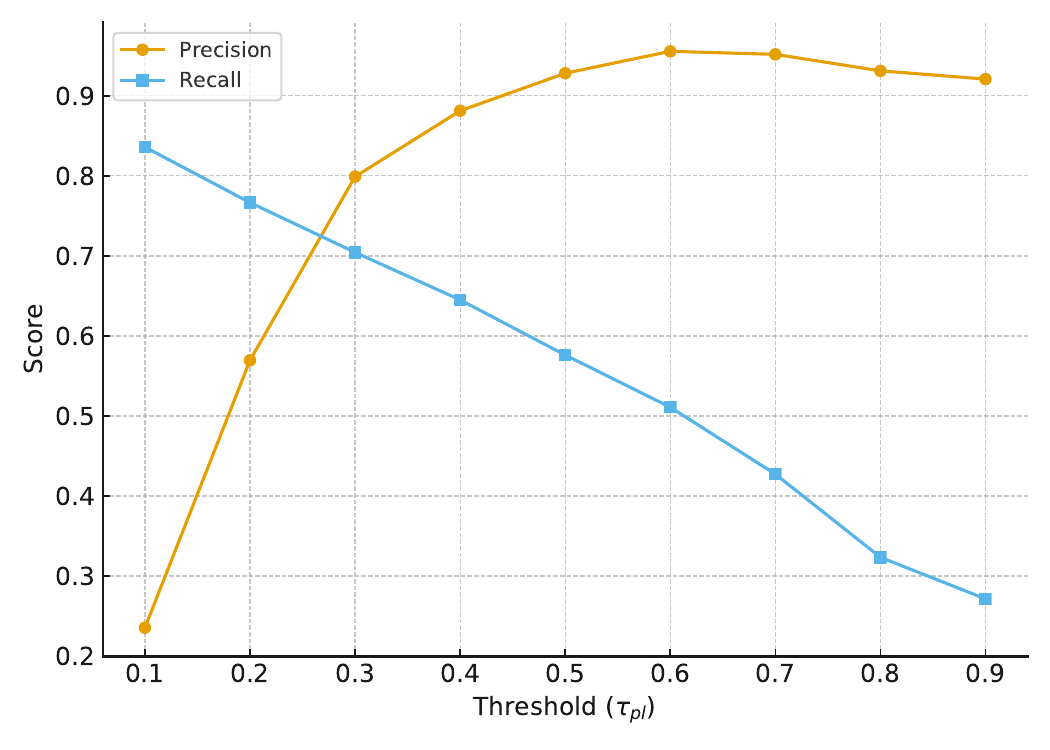}
\caption{Precision and recall of pseudo labels with respect to the confidence threshold $\tau_{\text{pl}}$.}
\label{fig:pl_threshold_tradeoff}
\end{figure}

Only the top $K$ detections are retained, and low-confidence predictions are filtered using a pseudo-label confidence threshold $\tau_{\text{pl}}$. The refined pseudo-label set is defined as
\[
\hat{\mathcal{Y}}_t = \{(\hat{b}_k, \hat{y}_k, \hat{s}_k) \in \mathcal{P}_t \mid \hat{s}_k \ge \tau_{\text{pl}}\},
\]
where $\hat{s}_k = \max_c p_{k,c}$ denotes the highest class confidence. The value of $\tau_{\text{pl}}$ determines a trade-off between precision and recall in the pseudo labels. A higher threshold increases precision by filtering out uncertain detections, however decreases recall, which means missing pseudo labels. Conversely, a lower threshold increases recall by retaining more true positives, but may introduce additional false positives that amplify noise during adaptation. As illustrated in Figure~\ref{fig:pl_threshold_tradeoff}, selecting an appropriate $\tau_{\text{pl}}$ is crucial to balancing stability and adaptability in the test-time learning process.

A non-maximum suppression with an IoU threshold $\tau_{\text{nms}}$ is then applied to eliminate redundant detections. The resulting pseudo labels $\hat{\mathcal{Y}}_t$ are used to supervise the student model $f_{\phi}$, which is initialized with the teacher’s parameters and optimized using the adaptation losses detailed in Section~\ref{sec:vertifocal_loss}. After each adaptation step, the teacher parameters are updated using an EMA of the student’s parameters:
\[
\theta_{t+1} \leftarrow \mu \, \theta_t + (1 - \mu) \, \phi_t,
\]
where $\mu$ is the EMA momentum coefficient. This update allows the teacher, which also serves as the deployed inference model, to remain continuously adapted to the evolving target domain while retaining the knowledge learned from the source data.

\subsection{Handling Pseudo-Label Noise with Varifocal Loss}
\label{sec:vertifocal_loss}
Pseudo labels generated by the teacher model inherently contain noise due to uncertain predictions under foggy conditions. Low visibility, degraded contrast, and domain shift often lead to inaccurate classification scores and imprecise localization of bounding boxes. Directly training the student model on such noisy pseudo labels can result in confirmation bias \cite{chen2022debiased, 9207304} and performance degradation. To mitigate this issue, we adopt the Varifocal loss~\cite{zhang2021varifocalnetiouawaredenseobject}, which adaptively weights the contribution of each sample based on its confidence and localization quality.

The Varifocal loss is formulated as:
\[
\mathcal{L}_{\text{vfl}} = -\sum_{i=1}^{N} \alpha_i \left( q_i (1 - p_i)^{\gamma} \log(p_i) + (1 - q_i) p_i^{\gamma} \log(1 - p_i) \right),
\]
where $p_i$ denotes the predicted probability, $q_i$ is the target confidence derived from the IoU between the predicted and pseudo-labeled boxes, and $\alpha_i$ and $\gamma$ are balancing and focusing parameters, respectively. Unlike hard thresholding, which discards low-confidence pseudo labels and their potentially useful information, the Varifocal loss retains all samples but adjusts their gradient contribution according to their confidence and localization accuracy. This allows the model to exploit informative low-confidence predictions while reducing the impact of unreliable ones.

By preserving pseudo labels and re-weighting them rather than removing them entirely, the Varifocal loss maintains richer supervision and achieves a better balance between learning stability and adaptability. Consequently, the adaptation process becomes more robust to pseudo-label noise and yields improved detection performance in foggy scenes compared to conventional focal loss.

\section{Experimental settings}
\label{sec:experimental_settings}

\subsection{Dataset}
\label{sec:dataset}
We conduct experiments on the clear-to-foggy domain adaptation scenario using the Cityscapes~\cite{Cordts2016Cityscapes} and Foggy Cityscapes~\cite{8917518} datasets. The labeled training set of Cityscapes is employed as the source domain, whereas the unlabeled training set of Foggy Cityscapes is utilized as the target domain during test-time adaptation. Following prior work, model performance is evaluated on the validation set of Foggy Cityscapes to measure its effectiveness and generalization capability under foggy-weather conditions.

\subsection{Implementation Details}
\label{sec:implementation}

To simulate the streaming nature of target data during test-time adaptation, the target images are processed sequentially with a batch size of one. The model is adapted for one epoch over the entire target dataset, ensuring that each image is seen exactly once in the adaptation phase. The learning rate is set to $1 \times 10^{-4}$. The pseudo-label confidence threshold $\tau_{\text{pl}}$ is set to 0.3, as discussed in Section~\ref{sec:pl_threshold}, and the NMS threshold $\tau_{\text{nms}}$ is fixed at 0.8 (Section \ref{sec:nms_threshold}). Based on common practice~\cite{tarvainen2018meanteachersbetterrole, moralesbrotons2024exponentialmovingaverageweights}, we set the EMA decay coefficient $\mu$ to 0.999. Following the default configuration of RF-DETR~\cite{rf-detr} for the maximum number of detected objects, we set $K = 300$ as the maximum number of pseudo labels retained after filtering.

Two types of data augmentation are applied to improve adaptation stability. For the teacher model, weak augmentation consists of light photometric and geometric transformations, including mild color jittering and random horizontal flipping. For the student model, a stronger augmentation is used to encourage robustness to appearance changes. It applies heavier color jittering, random grayscale conversion, Gaussian blur, and random erasing, combined with horizontal flipping.

All experiments are conducted on a single NVIDIA GeForce GTX 1080 Ti GPU.

\subsection{Evaluation Metric}
We evaluate performance using mAP, computed as the mean of class-wise AP values at an IoU threshold of 0.5. This metric jointly reflects detection accuracy and localization quality across all object categories.

\section{Experimental Results}
\label{sec:results}

\subsection{Results}
Table~\ref{tab:results} presents the detection performance under the Cityscapes $\rightarrow$ Foggy Cityscapes setting. Both the Source Only and Oracle models are based on the RF-DETR detector, where Source Only refers to the model trained solely on the clear-weather source domain without adaptation, and Oracle represents the model trained directly on labeled foggy data. The proposed FogTTA achieves the best overall performance across all classes, with an mAP of 57.6\%, outperforming state-of-the-art methods while maintaining real-time efficiency. Compared to the source-only baseline, FogTTA yields a +1.5\% mAP improvement and consistent gains across categories. These results confirm that online adaptation through pseudo-label refinement, Varifocal loss, and EMA-based teacher updates effectively mitigates domain shift and enhances robustness in foggy conditions.
\begin{table}[t]
\centering
\caption{Comparison of detection performance on the \textit{Cityscapes $\rightarrow$ Foggy Cityscapes} adaptation setting. $\dagger$ denotes Transformer-based object detection. TTA indicates test-time domain adaptation.}
\label{tab:results}
\resizebox{\linewidth}{!}{
\begin{tabular}{l|c|cccccccc|c}
\toprule
\textbf{Method} & \textbf{TTA} & \textbf{Person} & \textbf{Rider} & \textbf{Car} & \textbf{Truck} & \textbf{Bus} & \textbf{Train} & \textbf{Motor} & \textbf{Bicycle} & \textbf{mAP} \\
\midrule
MLFA \cite{10713112} & \checkmark & 38.0 & 50.9 & 18.5 & 32.3 & 14.4 & 21.0 & 31.9 & 22.5 & 28.7 \\
MemCLR \cite{VS_2023_WACV} & \checkmark & 32.1 & 41.4 & 43.5 & 21.4 & 33.1 & 11.5 & 25.5 & 32.9 & 29.8 \\
DDT \cite{ZHANG2025105740} & \checkmark & 32.0 & 41.1 & 47.9 & 25.1 & 33.8 & 28.5 & 26.0 & 35.2 & 33.7 \\
DA-Detect \cite{10030451} & $\times$ & 36.5 & 46.7 & 54.3 & 30.3 & 51.2 & 48.7 & 31.6 & 39.1 & 42.3 \\
DA-Detect (AdvGRL) \cite{10574400} & $\times$ & 36.5 & 49.1 & 55.8 & 30.0 & 48.7 & 51.3 & 33.3 & 41.6 & 43.4  \\ 
FRANCK \cite{11164935} $^{\dagger}$ & $\times$ & 48.1 & 49.3 & 60.6 & 33.9 & 48.2 & 36.9 & 34.0 & 47.9 & 44.9 \\
DDT with AEMA \cite{he2025dual} $^{\dagger}$ & $\times$ & 49.3 & 53.0 & 65.4 & 25.8 & 43.0 & 39.7 & 40.0 & 47.9 & 45.5 \\
\midrule
Source Only $^{\dagger}$ & \checkmark & 52.7 & 54.9 & 71.6 & 43.1 & 66.7 & 61.6 & 47.6 & 50.3 & 56.1 \\
\textbf{FogTTA (Ours)} $^{\dagger}$ & \checkmark & 53.8 & 58.0 & 72.6 & 45.2 & 68.3 & 63.1 & 48.1 & 52.0 & 57.6 \\
\midrule
Oracle $^{\dagger}$ & \checkmark & 57.1 & 58.8 & 78.6 & 47.0 & 76.4 & 67.2 & 48.5 & 52.6 & 60.8 \\
\bottomrule
\end{tabular}
}
\end{table}

\subsection{Qualitative Analysis}
\begin{figure*}[t]
\centering
\begin{minipage}{\linewidth}
\includegraphics[width=\textwidth]{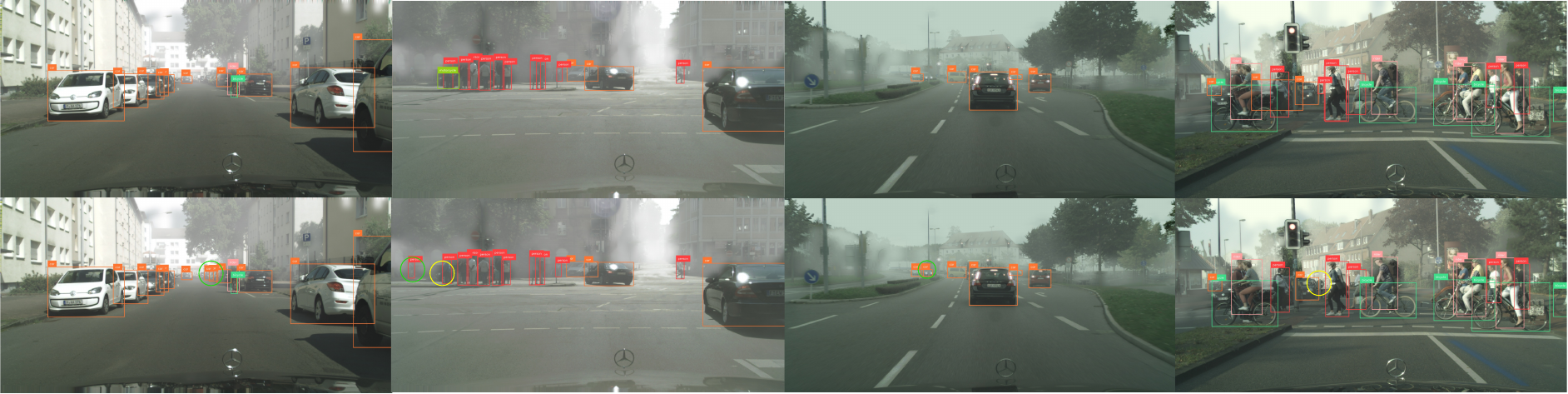}
\caption{Detection results before (top) and after (bottom) adaptation. Green circles show new correct detections, while yellow circles show false positives removed by FogTTA.}
\label{fig:qualitative_results}
\end{minipage}
\end{figure*}

Figure~\ref{fig:qualitative_results} shows detection results before (first row) and after (second row) adaptation. Green circles indicate new correct detections after adaptation, while yellow circles mark false detections removed by FogTTA, showing improved accuracy and robustness under foggy conditions.

\subsection{Ablation Study}
\label{sec:ablation}
The contribution of each component in the proposed framework is examined to assess its effect on adaptation performance. Table~\ref{tab:ablation} presents the results for different model variants, focusing on the influence of pseudo-label filtering and the Varifocal loss, which are both essential for achieving stable and reliable adaptation.

Removing pseudo-label filtering (FogTTA w/o Pseudo-Label Filtering) results in a clear decrease in accuracy. It shows the importance of discarding low-confidence detections to maintain high-quality supervision. Without this step, noisy pseudo labels with reduced precision are used during adaptation, which weakens the learning signal. Excluding the Varifocal loss (FogTTA w/o Varifocal Loss) also lowers performance, which highlights its effectiveness in mitigating residual noise by weighting samples according to their confidence and localization reliability. When both pseudo-label filtering and the Varifocal loss are removed, performance declines further, indicating that these two components complement each other in improving adaptation stability and overall detection accuracy.

\begin{table}[htbp]
\centering
\caption{Ablation study on the impact of pseudo-label filtering and Varifocal loss under the \textit{Cityscapes $\rightarrow$ Foggy Cityscapes} setting.}
\label{tab:ablation}
\resizebox{\linewidth}{!}{
\begin{tabular}{l|cccccccc|c}
\toprule
\textbf{Method} & \textbf{Person} & \textbf{Rider} & \textbf{Car} & \textbf{Truck} & \textbf{Bus} & \textbf{Train} & \textbf{Motor} & \textbf{Bicycle} & \textbf{mAP} \\
\midrule
FogTTA (Ours) & \textbf{53.8} & \textbf{58.0} & \textbf{72.6} & \textbf{45.2} & \textbf{68.3} & 63.1 & \textbf{48.1} & \textbf{52.0} & \textbf{57.6} \\
FogTTA w/o Pseudo-Label Filtering & 52.2 & 56.9 & 72.4 & 44.5 & 67.8 & 61.4 & 47.7 & 51.2 & 56.8 \\
FogTTA w/o Varifocal Loss & 52.1 & 57.5 & 71.0 & 43.1 & 67.2 & \textbf{63.8} & 46.9 & 51.1 & 56.6 \\
FogTTA w/o Pseudo-Label Filtering and Varifocal Loss & 52.1 & 55.9 & 70.1 & 44.5 & 68.9 & 62.3 & 45.8 & 50.6 & 56.3 \\
\bottomrule
\end{tabular}
}
\end{table}

\subsection{Sensitivity Analysis}
\subsubsection{Effect of $\tau_{\text{pl}}$}
\label{sec:pl_threshold}
\begin{figure}[htbp]
\centering
\includegraphics[width=0.5\textwidth]{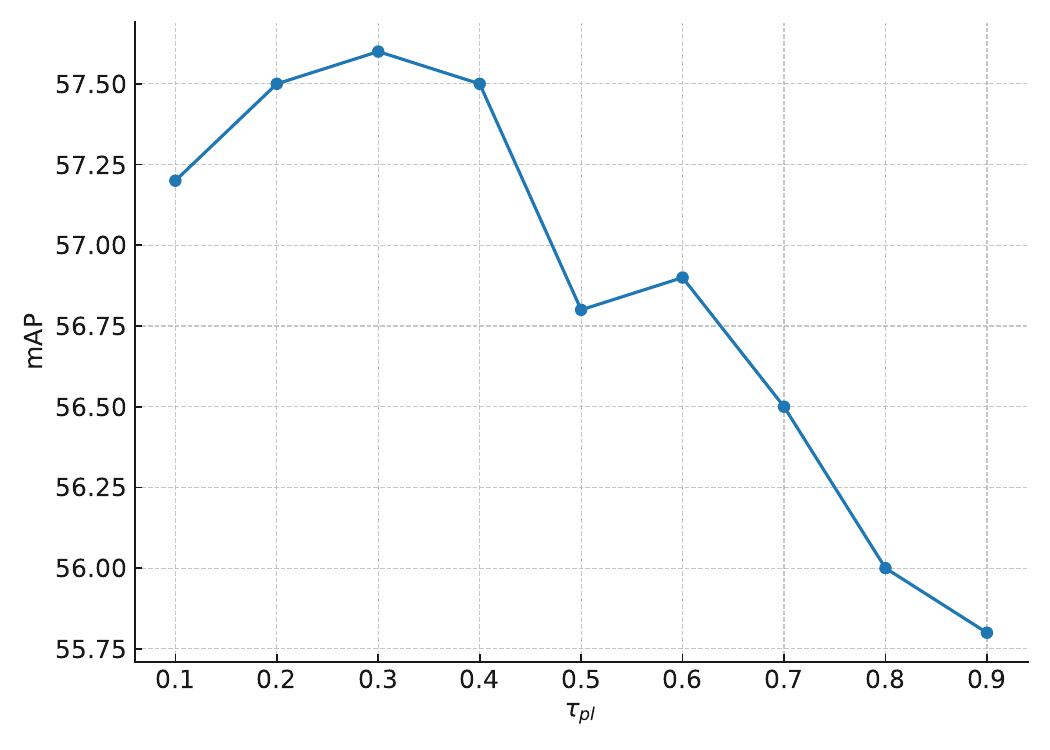}
\caption{Effect of $\tau_{\text{pl}}$ on mAP.}
\label{fig:tau}
\end{figure}

As illustrated in Figure~\ref{fig:tau}, the best overall performance is achieved when $\tau_{\text{pl}} = 0.3$. This observation aligns with the trend shown in Figure~\ref{fig:pl_threshold_tradeoff}, where a threshold around 0.3 provides a desirable balance between precision and recall. Lower values of $\tau_{\text{pl}}$ introduce excessive noise due to low-confidence predictions, while higher values filter out potentially informative pseudo labels, leading to reduced coverage. Setting $\tau_{\text{pl}}$ to 0.3 offers the best trade-off between stability and adaptability during test-time learning.

\subsubsection{Effect of $\tau_{\text{nms}}$}
\label{sec:nms_threshold}
The NMS threshold $\tau_{\text{nms}}$ controls the level of overlap allowed between detected bounding boxes and directly influences the trade-off between duplicate suppression and missed detections. Table~\ref{tab:nms_threshold} reports the results for different values of $\tau_{\text{nms}}$. As the threshold increases from 0.5 to 0.8, the mAP consistently improves, indicating that moderately higher overlap tolerance helps preserve valid detections that may otherwise be suppressed under dense or partially occluded conditions common in foggy scenes. When $\tau_{\text{nms}}$ exceeds 0.8, performance slightly declines due to increased redundancy from overlapping boxes. The best performance is achieved at $\tau_{\text{nms}} = 0.8$, which provides a balanced compromise between duplicate suppression and recall preservation.

\begin{table}[htbp]
\centering
\caption{Detection performance with different NMS thresholds $\tau_{\text{nms}}$.}
\label{tab:nms_threshold}
\resizebox{0.6\linewidth}{!}{
\begin{tabular}{l|cccccccc|c}
\toprule
\textbf{$\tau_{\text{nms}}$} & \textbf{Person} & \textbf{Rider} & \textbf{Car} & \textbf{Truck} & \textbf{Bus} & \textbf{Train} & \textbf{Motor} & \textbf{Bicycle} & \textbf{mAP} \\
\midrule
0.5 & 52.9 & 57.7 & 71.8 & 44.4 & 67.6 & 62.6 & 47.7 & 51.6 & 57.0 \\
0.6 & 52.0 & 57.3 & 71.5 & 44.3 & 67.9 & 64.6 & 48.4 & 52.0 & 57.2 \\
0.7 & 53.0 & 57.5 & 71.7 & 45.1 & 68.0 & 62.4 & 48.7 & 51.9 & 57.3 \\
0.8 & 53.8 & 58.0 & 72.6 & 45.2 & 68.3 & 63.1 & 48.1 & 52.0 & \textbf{57.6} \\
0.9 & 53.1 & 58.1 & 72.4 & 44.5 & 69.1 & 62.9 & 48.3 & 51.6 & 57.5 \\
\bottomrule
\end{tabular}
}
\end{table}

\section{Conclusion}
\label{sec:conclusion}
In this paper, we presented FogTTA, an online TTA framework designed to enhance the robustness of Transformer-based object detectors under foggy weather conditions. Building upon RF-DETR, FogTTA enables online adaptation without accessing source data by combining pseudo-label refinement, Varifocal loss, and EMA-based teacher updates. Experimental results demonstrated that FogTTA consistently outperforms all baseline methods in detection accuracy and stability compared to the source-only model, confirming the effectiveness of our approach for real-world autonomous driving scenarios.
For future work, we plan to extend FogTTA to broader adaptation scenarios, including cross-dataset settings and simulation-to-real transfer, to evaluate its generalization under more diverse domain shifts. We also aim to explore additional adaptation cues, such as temporal information and uncertainty estimation, to further improve robustness in challenging driving environments.

\section*{Acknowledgements}
This research is partially funded by the NSERC CREATE TrustCAV program and the NSERC Discovery Grant program, and in part by the NSERC Canada Research Chair (CRC) Program.

The authors used AI-assisted tools solely for language polishing of portions of this manuscript.

\printbibliography[heading=subbibintoc]

\end{document}